\pdfoutput=1

\documentclass[11pt]{article}

\usepackage[table]{xcolor}

\usepackage[preprint]{acl}

\usepackage{times}
\usepackage{latexsym}

\usepackage[T1]{fontenc}

\usepackage[utf8]{inputenc}

\usepackage{microtype}

\usepackage{inconsolata}

\usepackage{graphicx}

\usepackage{booktabs}
\usepackage{xspace}
\usepackage{microtype}
\usepackage{amsfonts}
\usepackage{amsmath}
\usepackage{amssymb}
\usepackage{amsthm}
\usepackage{booktabs}
\usepackage{dsfont}
\usepackage{paralist}
\usepackage{times}
\usepackage{xspace}
\usepackage{multirow}
\usepackage{graphicx}
\usepackage{float} 
\usepackage{lipsum}
\usepackage{appendix}
\usepackage{pifont}
\usepackage{makecell}
\usepackage{caption}
\usepackage{latexsym}
\usepackage{cleveref}
\usepackage{subcaption}
\usepackage{placeins}
\usepackage{upgreek}
\usepackage[inline]{enumitem}

\usepackage[export]{adjustbox}

\crefformat{section}{\S#2#1#3}
\crefformat{figure}{Fig.~#2#1#3}
\crefformat{table}{Tab.~#2#1#3}
\crefformat{appendix}{\S#2#1#3}

\renewcommand\paragraph[1]{
\vspace{0.15cm}
\noindent 
\textbf{#1}
}

\newcommand{\ra}[1]{\renewcommand{\arraystretch}{#1}}
\definecolor{lightgray}{gray}{0.95}
\usepackage{soul}

\usepackage{amsmath}
\usepackage{textcomp}
\usepackage{gensymb}

\newif\ifcomments
\commentstrue
\ifcomments
    \providecommand\uri[1]{[\textcolor{blue}{Uri: {#1}}]}

    \def \ifempty#1{\def\temp{#1} \ifx\temp\empty }

\else
    \providecommand{\uri}[1]{}
\fi

\title{\benchmarkname: A Challenging Benchmark for Complex Claim Verification}

 \author{Alon Jacovi$^1$ \ Moran Ambar$^1$ \ Eyal Ben-David$^1$ \ Uri Shaham$^1$ \\ [5px] \textbf{Amir Feder}$^1$ \ \textbf{Mor Geva}$^{1,2}$ \ \textbf{Dror Marcus}$^1$ \ \textbf{Avi Caciularu}$^1$ \\ [10px]
      $^1$Google Research  \quad  $^2$Tel Aviv University \\
        \tt alonjacovi@google.com}

\newcommand{\benchmarkname}{\textit{CoverBench}\xspace}

\begin{document}
\maketitle
\begin{abstract}


There is a growing line of research on verifying the correctness of language models' outputs. At the same time, LMs are being used to tackle complex queries that require reasoning. We introduce \benchmarkname, a challenging benchmark focused on verifying LM outputs in complex reasoning settings. Datasets that can be used for this purpose are often designed for other complex reasoning tasks (e.g., QA) targeting specific use-cases (e.g., financial tables), requiring transformations, negative sampling and selection of hard examples to collect such a benchmark. \benchmarkname provides a diversified evaluation for complex claim verification in a variety of domains, types of reasoning, relatively long inputs, and a variety of standardizations, such as multiple representations for tables where available, and a consistent schema. We manually vet the data for quality to ensure low levels of label noise. Finally, we report a variety of competitive baseline results to show \benchmarkname is challenging and has very significant headroom. 
The data is available at \url{https://huggingface.co/datasets/google/coverbench}.

\end{abstract}

\section{Introduction}

Recent work has focused on measuring various properties of language models' outputs~\cite{Leiter2022TowardsEE,Golovneva2022ROSCOEAS}. One established property to measure is the correctness of generated text, against its context~\cite{Tang2024MiniCheckEF} or external sources~\cite[e.g., fact-checking,][]{Pan2023FactCheckingCC,Chen2023FELMBF}.
This means, for example, to verify whether a summary correctly refers to its source document~\cite{Bishop2023LongDocFACTScoreET,krishna-etal-2023-longeval}, or that some logical inference has been made correctly based on claims that can be verified~\cite{jacovi2024chainofthought}.
We refer to this as \textit{claim verification}, where the claim is a falsifiable statement to be verified against a given grounding context~\cite{Honovich2022TRUERF}. It can be considered a reduction from NLI~\cite{DBLP:conf/mlcw/DaganGM05,Bowman2015ALA} or AIS~\cite{Rashkin2021MeasuringAI}.

\begin{figure}[t]
\setlength{\belowcaptionskip}{-10pt}
\centering
\resizebox{0.99\linewidth}{!}{
\includegraphics[valign=t]{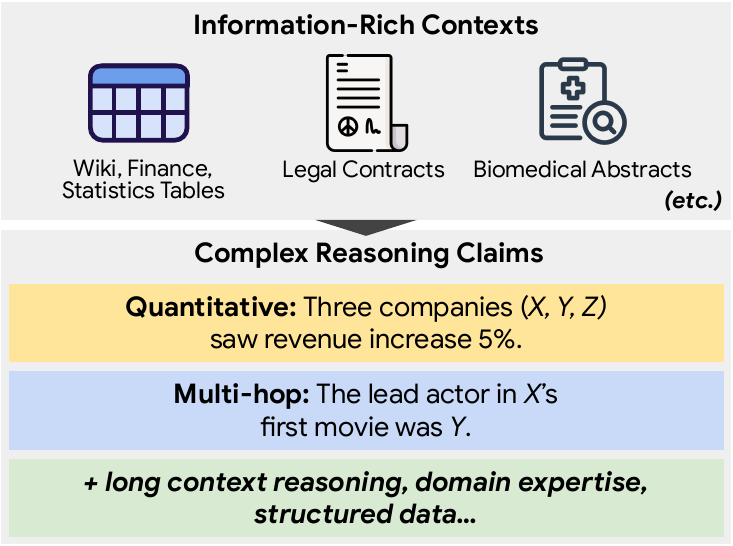}
}
\caption{\benchmarkname contains true and false claims that require implicit complex reasoning to verify in a variety of domains and settings.
}
\label{fig:teaser}
\end{figure}

\begin{table*}[t]\centering
\setlength\belowcaptionskip{-8px}
\setlength\tabcolsep{3pt}
\scriptsize
\rowcolors{1}{}{lightgray} \ra{1.2}
\resizebox{0.99\linewidth}{!}{
\begin{tabular}{
l
l
l
l
}\toprule
\textbf{Dataset} & \textbf{Domain} & \textbf{Task} & \textbf{Sources of Complexity}  \\\midrule
\textbf{\textit{FinQA}}~\cite{chen2021finqa} & Finance & QA & Quantitative, multi-step, tables \\
\textbf{\textit{QRData}}~\cite{liu2024llms} & Statistics & QA & Long context, quantitative, multi-step, domain expertise, tables \\
\textbf{\textit{TabFact}}~\cite{2019TabFactA} & Wikipedia & Verification & Multi-step, tables \\
\textbf{\textit{MultiHiertt}}~\cite{zhao-etal-2022-multihiertt} & Finance & QA & Long context, quantitative, multi-step, tables \\
\textbf{\textit{HybridQA}}~\cite{chen2020hybridqa} & Wikipedia & QA & Very long context, tables \\
\textbf{\textit{ContractNLI}}~\cite{chen2020hybridqa} & Legal & NLI & Long context, domain expertise \\
\textbf{\textit{PubMedQA}}~\cite{jin2019pubmedqa} & Biomedical & QA & Domain expertise \\
\textbf{\textit{TACT}}~\cite{caciularu2024TACT} & Various & QA & Quantitative, multi-step, tables \\
\textbf{\textit{Feverous}}~\cite{Aly21Feverous} & Wikipedia & Verification & Multi-step, quantitative, tables \\
\bottomrule
\end{tabular}
}
\caption{An overview of the datasets used in \benchmarkname (\Cref{subsec:variety}).}\label{tab:datasets}
\end{table*}


In this work, we focus on \textit{complex} claims. 
Naturally, as LMs are used frequently to solve complex queries~\cite{Suzgun2022ChallengingBT}, their verifying their outputs' correctness may require multiple hops of reasoning~\cite{Geva2021DidAU}, quantitative reasoning~\cite{Lewkowycz2022SolvingQR}, domain expertise~\cite{magesh2024hallucinationfree}, and so on, based on the reasoning level required in the original query.
Are general open-ended complex reasoning tasks and complex claim verification equivalent? We argue that there is a difference:
By focusing on a binary classification of given statements in a well-defined context, \textit{verifiers can---and should---be held to a higher standard compared to the source task}. We propose such a standard in this work.

To evaluate complex claim verification solutions, we collect \benchmarkname---a benchmark for this task---by leveraging a diverse set of nine datasets (\Cref{sec:scope}) across different settings that require complex reasoning (\Cref{fig:teaser}). The benchmark targets a variety of language domains (Wikipedia, finance, biomedical, legal, statistics, and others), sources of complexity (structured data, quantitative reasoning, multi-step reasoning, domain expertise, reasoning over long context), and difficulty and quality (via various filtering steps, both manual and automatic).

To build \benchmarkname (\Cref{sec:conversion}), we convert all tasks to a unified format with declarative claims, metadata about the required type of reasoning, and parsing and standardization of all table representations (we use \verb+HTML+, \verb+JSON+, and \verb+Markdown+). We additionally carefully sample false claims and challenging examples: In particular, challenging examples are selected through leveraging metadata and model-based selection. 
And similarly to TinyBenchmarks~\cite{polo2024tinybenchmarks}, we select an efficient subset of examples that will be the most representative for model ranking. 

\benchmarkname contains 733 examples of rich grounding contexts and complex claims based on them with correctness labels. The contexts are lengthy, with an average of 3,500 tokens. Many of the evaluated models, including recent competitive LMs, achieve performance near the random baseline, despite manual vetting we have done to ensure that the tasks are solvable. The best models achieve below a 70 Macro-F1 score, while smaller models---7--13 billion parameter LMs---achieve performance at the random baseline level. These results show a significant headroom for the task. The data is available at \url{https://huggingface.co/datasets/google/coverbench}.

\section{Benchmark Scope} \label{sec:scope}

For our new benchmark about complex claim verification, we prioritize \textit{variety} and \textit{difficulty}.
This section explains the facet of variety.

\label{subsec:variety}

\paragraph{Domains.} We aim to include a variety of language domains. In particular: The financial, Wikipedia, biomedical, and legal domains are well-represented in datasets for complex reasoning. A small quantity of examples in a large variety of other domains, such as statistical inference and literature, were included as well, within their limited availability.

\paragraph{Sources of complexity.} Complex reasoning can colloquially refer to many different tasks in practice. In the scope of this paper, we consider complex reasoning to include:
\begin{enumerate*}[label=(\arabic*)]
    \item Reasoning over structured data (in particular, tables);
    \item Reasoning over a long context;
    \item Quantitative reasoning (calculation, aggregation, counting, and so on);
    \item Reasoning requiring domain expertise;
    \item Multi-hop reasoning (inter-dependent steps of reasoning).
\end{enumerate*}
Importantly, we are interested not only in examples that exhibit any source of complexity, but specifically \textit{many unique combinations} of them.

\label{subsec:datasets}

\paragraph{Datasets.} In \Cref{tab:datasets} we describe the datasets in \benchmarkname. Some contexts with tables (e.g., in \textit{MultiHiertt}) contain a mix of table and text. Where possible, we leverage metadata to select the examples that require reasoning (e.g., as \textit{FinQA} contains the answers' calculation, we can select examples with multiple steps). Two originally-included datasets, \textit{SciTab}~\cite{lu2023scitab} and \textit{REVEAL}~\cite{jacovi2024chainofthought}, were excluded during a manual inspection process described in~\Cref{sec:conversion}. In \textit{Feverous}, different examples can exhibit different sources of complexity, and we select examples that exhibit at least one. Extended details are available in \Cref{app:benchmark-collection}.

\section{Constructing \benchmarkname} \label{sec:conversion}


We describe the process of collecting and building our benchmark. This involves conversion to a unified schema, negative sampling with seed models, and careful selection of informative examples.

\subsection{Conversion to the Schema}

Many of the datasets described in \Cref{subsec:datasets} require nuanced transformation into the claim verification setting. Below are relevant details that require attention in this transformation.

\paragraph{Schema.}
Each example in \benchmarkname contains the following:
The grounding \textit{context}; a falsifiable \textit{claim} in the form of a declarative statement; a binary entailment \textit{label}; the language \textit{domain} of the instance; and the \textit{sources of complexity} that make this instance require complex reasoning (see \Cref{sec:scope}).

\paragraph{QA pairs to declarative statements.} Claims which come in the format of a QA pair (e.g., ``Q: What is the capital of France? A: Paris.'') were converted to declarative form (``The capital of France is Paris.'') through both manual annotation and a prompted LM (all generated statements were manually verified; details in \cref{app:benchmark-collection}).

\paragraph{Representing structure.} The various table formats across datasets were parsed into a standard format, which we represented in one of three text formats, chosen at random: \verb+HTML+, \verb+JSON+, or \verb+Markdown+.

\subsection{Sampling Negative Examples}

\textit{TabFact}, \textit{ContractNLI}, \textit{PubMedQA}, and \textit{Feverous} are datasets that contain both true and false claims (in the case of \textit{PubMedQA}, since answers are binary, the inverse answer can also be used.). 

\textit{FinQA}, \textit{QRData}, \textit{MultiHiertt}, \textit{HybridQA} and \textit{TACT} are QA datasets---and so they contain only positive examples as QA pairs. For these tasks we require a method to generate negative examples---claims not entailed by the context. The negative examples should be difficult, which precludes simpler heuristics for negative sampling~\cite{Li2019SamplingMA}.

Given the original QA format, a simple method to derive difficult negative cases is to use the question to generate models' answers. The answers are compared to the gold answer, and if they are wrong, the new QA pair is selected as a negative example. These negative answers represent real model errors, so they are likely difficult for models to verify. 

We used the following three seed models: \textit{GPT-4o}~\cite{openai2024gpt4}, \textit{gemma-1.1-7b-1.1-it}~\cite{gemmateam2024gemma}, and \textit{Mixtral-8x7b-Instruct}~\cite{jiang2024mixtral}. After extracting the classification from the models' full answer, we compared the gold and model answers in two ways: Lowercased exact match (percentages, currencies other numbers were normalized to digits), and using a prompted LM judge (we used \textit{Mixtral-8x7b}, and manually verified perfect accuracy for this simple task on a representative sample of 100 cases). The generated answer was considered incorrect if it was judged as different for both heuristics.

\subsection{Example Selection} \label{subsec:subset-sampling}

Following the methodology above, we derived a total of roughly 7,000 examples. From this larger set, we selected a subset of examples with three goals:
\begin{enumerate*}[label=(\Roman*)]
    \item Most difficult examples, without introducing label noise bias.
    \item Examples that are  least likely to suffer from memorization via data contamination.
    \item Examples that are  most indicative of the difference between models' capabilities.
\end{enumerate*}

To clarify goal (I): It is expected that all datasets have annotation errors~\cite{Klie2022AnnotationED}. E.g., if we simply select examples of model mistakes, while this will select difficult examples~\cite{10.1162/coli_a_00464}, it would also select a greater ratio of incorrectly-labeled examples. Thus, we cannot select examples of model mistakes deliberately.

Goals (I) and (II) can be targeted by selecting examples that are incorrectly predicted by models in the \textit{claim-only} baseline setting. In this setting, only the claim is provided to the model without its grounding context, so the example is intractable, which avoids the incorrect label bias. If the model is correct---it is via a random guess, some shallow heuristic~\cite{McCoy2019RightFT}, or data contamination~\cite{Deng2023InvestigatingDC}.\footnote{We cannot guarantee against data contamination within the scope of this paper, since many models use hidden training data. Nevertheless, we rely on overall similarities between models and their similar internet-derived training data. We refer to the individual model developers to index \benchmarkname and check for contamination when reporting evaluations on it.} By choosing examples where at least two models were incorrect in this baseline, we reduce the possibility of discarding correctness via a random guess.

Goals (II) and (III) can be targeted by selecting model disagreements. Since on any disagreement one model was correct and one was incorrect, we can select examples---without using gold labels---which demonstrably differentiate between models. 

We made both selections by using two comparable-performance seed models: \textit{Mixtral-8x7b-Instruct} and \textit{Starling-LM-7B-beta-ExPO} \cite{starling2023}.
Performance on the final subset was 3 to 7 Macro-F1 points lower than on a random subset in our testing (\Cref{app:experiments}).

\begin{figure}[t]
\setlength{\belowcaptionskip}{-10pt}
\centering
\resizebox{0.9\linewidth}{!}{
\includegraphics[valign=t]{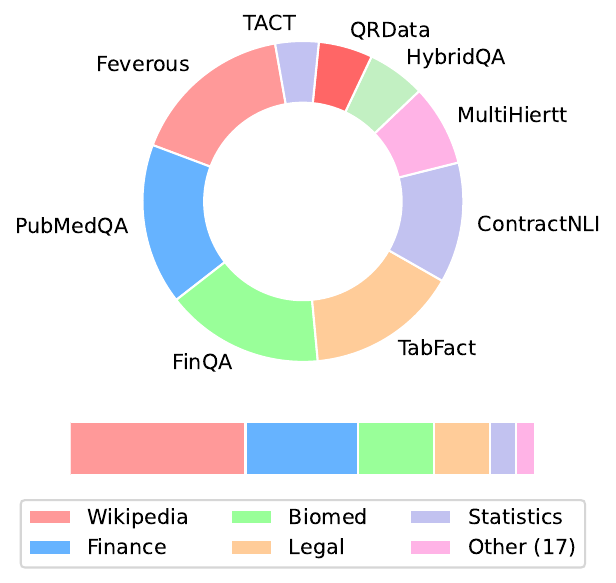}
}
\caption{Distribution of the source datasets and the text domains in \benchmarkname.
}
\label{fig:stats}
\end{figure}

\subsection{Vetting}

Due to the various steps described above, we require a phase of manual inspection to check for solvability of the derived data and diagnose possible issues. We selected 10 examples at random from each dataset for a manual vetting phase. The datasets had between 0 and 1 incorrect labels, indicating a rate of incorrect labels comparable with the academic datasets that the data is based on.

During this phase, two datasets originally selected were omitted: \textit{SciTab}~\cite{lu2023scitab} was omitted due to a high level of label noise, and \textit{REVEAL}~\cite{jacovi2024chainofthought} was omitted due to a loss of difficulty (notably, this loss of difficulty was an artifact of the conversion, and not present in the original data). \textit{PubMedQA} also exhibited a loss of difficulty during the conversion process via a correlation between negation in the claim and a non-entailment label. However, this issue was neutralized with the \textit{claim-only sampling} step  (\Cref{subsec:subset-sampling}).

\subsection{The \benchmarkname Benchmark}



The final challenge set contains 733 examples. \Cref{fig:stats} shows the distribution of the source datasets and domains. The examples have 3,500 tokens on average via the \textit{Mixtral-8x7b-Instruct} tokenizer.
The overall label distribution is balanced at 45:55 towards the positive class (58:42 to 44:56 per source dataset). More details are in \Cref{app:benchmark-collection}.


\begin{table}[t]\centering
\setlength\belowcaptionskip{-8px}
\setlength\tabcolsep{3pt}
\scriptsize
\rowcolors{1}{}{lightgray} \ra{1.2}
\resizebox{0.95\linewidth}{!}{
\begin{tabular}{lcc}\toprule
\textbf{Model} &\textbf{0-shot} &\textbf{0-shot CoT} \\\midrule
NLI-Entailment-Verifier-xxl &43.7 & --- \\
T5-11b-TrueTeacher\&ANLI &48.2 & --- \\
$*$ Gemma-1.1-7b-it &43.4 &46.1 \\
$*$ Mixtral-8x7B-Instruct &45.3 &49.0 \\
Llama-2-13b-Chat &48.3 &48.7 \\
Qwen1.5-14B-Chat &49.0 &50.8 \\
$*$ Starling-LM-7B-beta-ExPO &47.7 &52.1 \\
Llama-2-70b-Chat &50.3 &52.6 \\
Yi-1.5-34B-Chat &51.0 &54.2 \\
Gemini 1.5 Flash &54.4 &54.5 \\
Qwen2-72B-Instruct &54.3 &57.0 \\
Gemini 1.5 Pro (05/2024) &59.9 &62.1 \\
$*$ GPT-4o (10/2024) &62.7 & 66.5 \\
Claude 3.5 Sonnet (10/2024) & 67.2 & 68.9 \\
\bottomrule
\end{tabular}
}
\caption{Macro-F1 performance on \benchmarkname. 50 is the random baseline threshold, and below 50 implies a class bias, where 0 is the majority baseline. ($*$) denotes seed models for sampling---we include their results for completeness, but note that they are unreliable.}\label{tab:f1-results}
\end{table}

\section{Experimental Setup and Results}

In this section, we describe the experiments and results we made to evaluate the difficulty of \benchmarkname. The Macro-F1 results are in \Cref{tab:f1-results}. Overall, we see that a variety of competitive models struggle on this task. Fine-grained details of the experiment setting and our approach are available in \Cref{app:experiments}.


\paragraph{Baseline details.} The baselines use prompted LMs~\cite{gemmateam2024gemma,ai2024yi,qwen1.5,qwen2,reid2024gemini,touvron2023llama,anthropic_claude_2024} and off-the-shelf NLI classifiers~\cite{gekhman2023trueteacher,sanyal2024machines}. The LMs were prompted in 0-shot and 0-shot Chain-of-Thought~\cite{wei2023chainofthought} formats, after extensive ``prompt engineering'' using prompts from previous works and trial-and-error. While we made significant attempts with few-shot settings, none succeeded, likely due to the fact that the examples in \benchmarkname are often long (we have tested both simplified and full-length demonstrations), as few-shot is known to struggle in long-context~\cite{Jiang2023LongLLMLinguaAA,cattan2024fewshotworklongcontextrecycling}. For the NLI classifiers, contradiction and neutral were considered as not entailed. In cases where the context length exceeded the model's context window, if the model did not include an extrapolation technique, the context was trimmed from its beginning.





\section{Conclusions}

We collect a new claim verification benchmark specifically targeting difficulty and complexity of reasoning, towards the goal of developing not only models with good complex reasoning, but \textit{better  verification} to classify when a generated claim is false. 
\benchmarkname involves a wide variety of domains and sources of complexity, has long inputs with thousands of tokens on average, and is relatively efficient in size. The benchmark provides a significant challenge to current models, and can serve as a groundwork for future work in the area. 

\section*{Limitations}

\paragraph{Domain-specific LMs.} In this work, we focused on validating the difficulty of the benchmark based on readily-available off-the-shelf LMs. Some of the tasks will likely be better addressed by LMs that use specialized tools or specialized prompting techniques~\cite{Kim2024HuskyAU}, LMs that were specifically trained for a specific domain such as finance~\cite{Wu2023BloombergGPTAL}, and so on \cite{pan2023factcheckingcomplexclaimsprogramguided}. Our goal is to measure general ability in diverse settings, but specialized use-cases may benefit from investigating specialized models on a the relevant subset of \benchmarkname.

\paragraph{Data contamination disclaimer.} As mentioned in the paper, despite some steps we took against this (such as generating negative examples, converting QA to declarative statements, converting tables to different formats than those in the original datasets, and sampling less-memorized examples), the problem of data contamination limits the ability of evaluation datasets currently. Evaluators using the data should take care to check for the individual examples' presence in their models' training data, if they have access to it, and otherwise they should not rely on the metrics for critical decisions about models whose training data is unknown. The role of the evaluations in this work is strictly to illustrate a robust general property of the data as a difficult evaluation set with large headroom.


\bibliography{anthology,custom}

\clearpage

\appendix

\section{Benchmark Collection} \label{app:benchmark-collection}



\begin{figure}[t]
\centering
\resizebox{0.9\linewidth}{!}{
\includegraphics[valign=t]{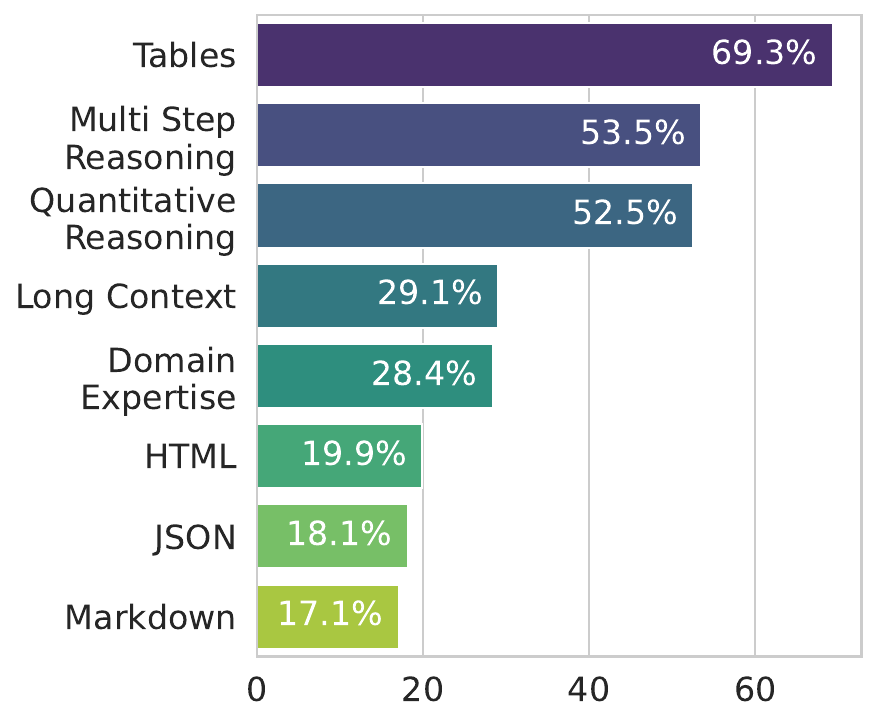}
}
\caption{Distribution of the sources of complexity in \benchmarkname. Long context in this figure refers to examples with over 3,000 tokens with the \textit{Mixtral-8x7b-Instruct} tokenizer. 
}
\label{fig:complexity-dist}
\end{figure}

Benchmark statistics are given in \Cref{tab:app-statistics,fig:complexity-dist}. Examples from each dataset in the benchmark are in \Cref{tab:datasets-long}.

The 17 domains which are included under ``Other'' in \Cref{fig:stats}, and are represented by very few examples in the dataset (a total of 31 examples), are: business, media, culinary, emergency, environment, fashion, medicine, archaeology, ethics, real-estate, politics, technology, biology, transportation, retail, education, and architecture. These domains are all from the TACT dataset which has a wide variety of domains.

\subsection{Datasets}

Below we describe each dataset and what it was used for. The public test sets were used when available, and otherwise, the public dev sets. Each instance of \benchmarkname contains the precise ID of where the instance came from with respect to the source datasets.

\paragraph{FinQA~\cite{chen2021finqa}} includes QA pairs that require quantitative reasoning (with a numerical answer) over a medium-length text and a table extracted from a financial report. Since the data includes the entire calculation behind the answer, we selected QA pairs that require at least two steps of calculations.

\paragraph{QRData~\cite{liu2024llms}} contains QA pairs that require statistical inference or causal inference over a large table. We selected QA pairs that require statistics, as difficult examples of quantitative reasoning. 

\paragraph{TabFact~\cite{2019TabFactA}} contains Wikipedia tables and declarative statements based on the tables, some of which require reasoning over multiple cells in the table (e.g., aggregation or \textit{arg-max}). We manually selected examples that require reasoning over multiple cells.

\paragraph{MultiHiertt~\cite{zhao-etal-2022-multihiertt}} similarly to FinQA includes QA pairs that require multi-step quantitative reasoning, with a numerical answer, in the finance domain. The contexts in MultiHiertt are long, with multiple tables (some of them hierarchical) interspersed throughout a long document. 

\paragraph{HybridQA~\cite{chen2020hybridqa}} includes multi-hop QA pairs over a mix of text and a table in the Wikipedia domain.

\paragraph{ContractNLI~\cite{koreeda-manning-2021-contractnli-dataset}} includes natural language inference examples, where the premises are non-disclosure agreements, and the hypotheses are a set of 17 standardized conclusions. In the majority of cases, domain expertise is required to derive the correct label.

\paragraph{PubMedQA~\cite{jin2019pubmedqa}} includes binary (true-false) QA pairs over PubMed abstracts. A large quantity of examples require reasoning over multiple sentences to answer, and all require domain expertise.

\paragraph{TACT~\cite{caciularu2024TACT}} includes text-table alignments alongside QA pairs that require multiple complex calculations, formalized as chains of table-manipulation queries, in a large variety of domains. 

\paragraph{Feverous~\cite{Aly21Feverous}} includes claims over Wikipedia documents, many of which include tables and quantitative reasoning or multi-step reasoning. We selected the examples that require tables, and either quantitative reasoning or multi-step reasoning.

\begin{table}[t]\centering
\setlength\tabcolsep{3pt}
\scriptsize
\rowcolors{1}{}{lightgray} \ra{1.2}
\resizebox{0.9\linewidth}{!}{
\begin{tabular}{lrcc}\toprule
Dataset &Tokens &Label Ratio (+) & Size \\\midrule
{\textit{ContractNLI}} &2,572 &42.7\% & 12.1\% \\
{\textit{Feverous}} &5,204 &56.2\% & 16.5\%\\
{\textit{FinQA}} &1,252 &57.3\% & 16.0\%\\
{\textit{HybridQA}} &19,297 &55.8\% & 5.9\%\\
{\textit{MultiHiertt}} &4,829 &58.3\% & 8.2\%\\
{\textit{PubMedQA}} &360 &56.3\%& 16.2\% \\
{\textit{QRData}} &5,512 &57.5\% & 5.5\%\\
{\textit{TACT}} &314 &56.3\% & 4.4\%\\
{\textit{TabFact}} &1,239 &56.3\% & 15.3\%\\
\bottomrule
\end{tabular}
}
\caption{Statistics per source dataset.}\label{tab:app-statistics}
\end{table}

\subsection{Conversion to the Schema}

\paragraph{Conversion to declarative statements.} For generating declarative statements, we used multiple LMs, including Llama-2-7b and Mixtral-8x7b; all generated statements were manually verified. Where the generated statements were incorrect, the authors of this work manually wrote the declarative statements.

\paragraph{Tables.} 
Each dataset's tables were parsed using the dataset's associated source code when available, and otherwise, the parsing was written by us. All tables were parsed into pandas DataFrames~\cite{reback2020pandas}. The tables' text representations of \verb+HTML+, \verb+JSON+, and \verb+Markdown+ were derived using the default implementations in pandas (the \verb+JSON+ representation uses the ``records'' setting) and chosen at random for each example. Since some of the datasets use the same context for multiple different claims, the some contexts may appear multiple times in \benchmarkname, either with the same table representation or different ones. In the case of Feverous, due to the unique typesetting of the Wikipedia tables employed in that dataset, the tables were taken as-is.

\paragraph{Negative sampling.} The final distribution of negative examples by their answering model, after all selection and sampling steps, are: 48.8\% (\textit{Mixtral-8x7B-Instruct}), 28.8\% (\textit{gpt-4o}),  22.4\% (\textit{gemma-1.1-7b-it}).

\subsection{Prompts.}
The prompts we used for the sub-tasks in the conversion process are given below.

\vspace{0.15cm}
\noindent\textit{QA to Declarative Statement:}
\begin{quote}
    ``Edit the following question and answer into a declarative form.

For example, given the question "What is the population of Europe? Round to the nearest million." and answer "741,000,000", output "The population of Europe, rounded to the nearest million, is 741,000,000."

Question: \textit{[question]}

Answer: \textit{[answer]}

Declarative form:''
\end{quote}

\noindent\textit{Final Answer Extraction:}
\begin{quote}
    ``Given a question and a model's answer, extract the pure term answer from this text.
    
    For example if the answer to the question ``How much did the stocks rise in 2001?'' is ``The stocks in 2005 rose \$50 from 2001.'', generate ``\$50''. If the model didn't give a clear answer, output ``None''.
    
  Question: \textit{[question]}
  
  Model answer: \textit{[model answer]}
  
  Extracted final answer:''
\end{quote}

\noindent\textit{Gold Answer Comparison Check:}
\begin{quote}
    ``Given two terms, Term 1 and Term 2, your task is to compare the two terms and say whether they are equivalent or not.
    
  For example, ``12\%'' and ``0.12'' are equivalent, and ``2 thousand dollars'' is equivalent to ``\$2,000'', but different values, different units or different entities are not equivalent. Generate ``Yes'' or ``No''.
  
  Term 1: \textit{[model answer]}
  
  Term 2: \textit{[gold answer]}
  
  Equivalent:''
\end{quote}

\begin{table}[t]\centering
\setlength\belowcaptionskip{-8px}
\setlength\tabcolsep{3pt}
\scriptsize
\rowcolors{1}{}{lightgray} \ra{1.2}
\resizebox{0.99\linewidth}{!}{
\begin{tabular}{lcc}\toprule
\textbf{Model} &\textbf{Our sample} &\textbf{Random sample} \\\midrule
$*$Gemma-1.1-7b-it &46.1 &48.5 \\
$*$Starling-LM-7B-beta-ExPO &52.1 &58.2 \\
$*$Mixtral-8x7B-Instruct &49.0 &59.6 \\
Gemini 1.5 Flash &54.5 &57.8 \\
Yi-1.5-34B-Chat &54.2 &61.1 \\
Qwen2-72B-Instruct &57.0 &62.0 \\
Gemini 1.5 Pro &62.1 &68.8 \\
\bottomrule
\end{tabular}
}
\caption{Macro-F1 performance on \benchmarkname (0-shot CoT) for comparison between our selection of examples (\Cref{subsec:subset-sampling}) and a random sample of the same size. ($*$) denotes seed models for sampling---we include their results for completeness, but note that they are unreliable.}\label{tab:f1-sample-comparison}
\end{table}

\section{Experiment Details} \label{app:experiments}

\subsection{Implementations}

For the LM baselines, binary model decisions were parsed from model outputs under greedy decoding, according to the format below. If a model failed to comply with the prompt format, we additionally concatenated ``Final answer (correct or incorrect):'' to the end of the model's answer and re-prompted it to get the final answer for parsing.

The implementations and prompts were checked for soundness by seeing the gap between our measurements on a random subset compared to measurements in the literature, to observe that they are overall similar. The code to use the models was based on standard available HuggingFace~\cite{wolf2020huggingfaces} code for each model, alongside their associated tokenizers and chat templates. For Gemini 1.5 Flash and Pro, the API version was used. 

\subsection{Additional Results}

\Cref{tab:f1-sample-comparison} shows a comparison of results between a random selection of examples and our model-based selection.

\subsection{Prompts}

Below are the prompts we used in the baseline evaluations. We have tested various edits of the prompts on a random subset of examples which were not included in the final benchmark.


\vspace{0.15cm}
\noindent\textit{Zero-Shot:}
\begin{quote}
``Your task is to check if the Claim is correct according to the Evidence. Generate 'Correct' if the Claim is correct according to the Evidence, or 'Incorrect' if the claim is incorrect or cannot be verified.

Evidence: \textit{[context]}

Claim: \textit{[claim]}

Answer:''
\end{quote}

\noindent\textit{Zero-Shot with Chain-of-Thought:}
\begin{quote}
``Your task is to check if the Claim is correct according to the Evidence. Generate 'Correct' if the Claim is correct according to the Evidence, or 'Incorrect' if the claim is incorrect or cannot be verified.

Evidence: \textit{[context]}

Claim: \textit{[claim]}

Let's think step-by-step:''
\end{quote}

\begin{table*}[t]\centering
\setlength\belowcaptionskip{-8px}
\setlength\tabcolsep{3pt}
\scriptsize
\rowcolors{1}{}{lightgray} \ra{1.1}
\resizebox{0.99\linewidth}{!}{
\begin{tabular}{
>{\raggedright\arraybackslash}p{1.4cm}
>{\raggedright\arraybackslash}p{8.5cm}
>{\raggedright\arraybackslash}p{3.2cm}
}\toprule
\textbf{Dataset} & \textbf{Context (snippet)} & \textbf{Claim} \\\midrule
\textbf{\textit{FinQA}}& Paper Title: One-to-X analogical reasoning on word embeddings: a case for diachronic armed conflict prediction from news texts\newline
Table 3: Average diachronic performance\newline
|    | [EMPTY]   | [BOLD] Algorithm   |   [BOLD] Precision |   [BOLD] Recall | [BOLD] F1   |\newline
|---:|:----------|:-----------------|---------------:|--------------:|:----------|\newline
|  0 | Giga      | Baseline           |               0.19 |            0.51 | 0.28        |\newline
|  1 | Giga      | Threshold          |               0.46 |            0.41 | [BOLD] 0.41 | 
\textit{[...]}& For both Gigaword and NOW datasets (and the corresponding embeddings), using the cosine-based threshold increases recall and decreases precision (differences are statistically significant with t-test, p < 0.05).  \\
\textbf{\textit{QRData}} & The Stanford University Heart Transplant Study was conducted to determine whether an experimental heart transplant program increased lifespan. Each patient entering the program was designated an official heart transplant \textit{[...]} \newline
The data is in the CSV file heart\_transplant.csv. \newline
heart\_transplant.csv \newline
|     |   id |   acceptyear |   age | survived   |   survtime | prior   | transplant   |   wait | \newline
|----:|-----:|----------:|------:|:--------|--------:|:------|:----------|-----:|\newline
|   0 |   15 |           68 |    53 | dead       |          1 | no      | control      |    nan |\newline
|   1 |   38 |           70 |    41 | dead       |          5 | no      | treatment    |      5 |
\textit{[...]} & The difference in survival rate between the control and treatment groups can be estimated using a confidence interval constructed with the normal approximation. \\
\textbf{\textit{TabFact}} & \{"artist": \{"0": "ophiolatry", "1": "ophiolatry", "2": "black flame", "3": "tangorodrim", "4": "tangorodrim", "5": "triumfall"\}, "title": \{"0": "transmutation", \textit{[...]} 
& Black Flame's title was released five months and two days after Ophiolatry's 2008 title. \\
\textbf{\textit{MultiHiertt}} & HEWLETT PACKARD ENTERPRISE COMPANY AND SUBSIDIARIES Management’s Discussion and Analysis of Financial Condition and Results \textit{[...]} \newline
\#\# Table 0 \#\# \newline
<table><tr><td></td><td colspan="3">For the fiscal years ended October 31,</td>\textit{[...]}<td>2019</td><td>2018</td><td>2017</td></tr><tr><td></td><td colspan="3">Dollars in millions</td></tr><tr><td>Net revenue \textit{[...]} \newline
Fiscal 2019 compared with Fiscal 2018 Corporate Investments net revenue decreased by \$36 million, or 6.6\% (decreased 4.4\% on a constant currency bases), in \textit{[...]} \newline
\#\# Table 1 \#\# \newline
<table><tr><td> \textit{[...]}
& Two amounts for Credit Card (a) exceeded the average of Amount for Credit Card (a) in 2012.  \\
\textbf{\textit{HybridQA}}  & 1998 IAAF World Half Marathon Championships \newline
The 7th IAAF World Half Marathon Championships was held on September 27, 1998, in the city of Uster, Switzerland. A total of 236 athletes, 139 men and 97 women, from 54 countries took part. Detailed reports on the event and an appraisal of the results was given. Complete results were published. \newline
Team Results -- Men 's \newline
<table border="1" class="dataframe">
  <thead>
    <tr style="text-align: right;">
      <th></th>
      <th>Rank</th>
      <th>Country</th> \textit{[...]} & Ethiopia finished in 3:05:18 at the IAAF World Half Marathon Championships of 1998 southeast of the country. \\
\textbf{\textit{ContractNLI}}& DEPARTMENT OF HOMELAND SECURITY \newline
NON-DISCLOSURE AGREEMENT \newline
I, \_\_, an individual official, employee, consultant, or subcontractor  of or to \_\_ (the Authorized Entity), intending to be legally bound, hereby consent to the terms in this Agreement in consideration of my being granted conditional access to certain information, specified below, that is owned by, produced by, or in the possession of the United \textit{[...]} & Confidential Information shall only include technical information. \\
\textbf{\textit{PubMedQA}} & To examine whether government-funded, low-income vision care programs improve use of eye care services by low-income individuals in Canada. \newline
Cross-sectional survey. \newline
27,375 white respondents to the Canadian Community Health Survey (CCHS) Healthy Aging 2008/2009.
Government-funded, low-income vision care programs were reviewed. The amount of assistance provided was compared with professional fee schedules for general/routine eye examinations  \textit{[...]} & Government assistance may improve utilization of eye care services by low-income individuals. \\
\textbf{\textit{TACT}} & [\{"Job Title":"Registered Nurse (RN)","Job Description":"Full-Time position for caring for patients at Gentiva Home Hospice, Monday - Friday, daytime schedule, reports to Hospice, RN, Patient Care.","Location":"Bloomington, IL","Salary":"\$20.04 \textit{[...]} & The average salary described in the text is 27.52 per hour. \\ 
\textbf{\textit{Feverous}} & [H] Position: | [[Offensive\_tackle|Offensive tackle]]/[[Guard\_(gridiron\_football)|Guard]]\newline
[H] Personal information | [H] Personal information\newline
[H] Born: | (1947-08-30) August 30, 1947 (age 73)\newline
[[Ponca\_City,\_Oklahoma|Ponca City, Oklahoma]]\newline
[H] Height: | 6 ft 2 in (1.88 m)\newline
[H] Weight: | 262 lb (119 kg) \textit{[...]} & Jon Kolb ( played 200 games as a professional football player and he weighs 199lb). \\
\bottomrule
\end{tabular}
}
\caption{Examples from each of the source datasets used in \benchmarkname. The labels are true or false.}\label{tab:datasets-long}
\end{table*}

\end{document}